\pdfoutput=1
\documentclass{article}
\usepackage{spconf,amsmath,graphicx}

\usepackage{xcolor}
\usepackage{subcaption}
\usepackage{multirow}
\usepackage{multicol}
\usepackage{booktabs}
\usepackage{algpseudocode}
\usepackage{algorithm}
\usepackage{amssymb}
\usepackage{optidef}
\usepackage{url}
\usepackage{adjustbox}

\usepackage{paralist}
\usepackage{microtype}

\title{Confidence-based federated distillation for vision-based lane-centering}
\name{Yitao Chen$^{*}$, Dawei Chen$^{\dagger}$, Haoxin Wang$^{\ddagger}$, Kyungtae Han$^{\dagger}$, Ming Zhao$^{*}$\thanks{This work is partly supported by National Science Foundation awards CNS-1955593 and OAC-2126291.}}
\address{$^{*}$Arizona State University, $^{\dagger}$Toyota InfoTech Labs, $^{\ddagger}$Georgia State University}
\algtext*{EndWhile}
\algtext*{EndIf}
\algtext*{EndFor}
\algtext*{EndProcedure}

\newcommand{\yitao}[1]{\textcolor{black}{#1}}
\newcommand{\yt}[1]{\textcolor{black}{#1}}

\begin{document}
%
\maketitle
%


\begin{abstract}
A fundamental challenge of autonomous driving is maintaining the vehicle in the center of the lane by adjusting the steering angle. Recent advances leverage deep neural networks to predict steering decisions directly from images captured by the car cameras. Machine learning-based steering angle prediction needs to consider the vehicle's limitation in uploading large amounts of potentially private data for model training. 
Federated learning can address these constraints by enabling multiple vehicles to collaboratively train a global model without sharing their private data, but it is difficult to achieve good accuracy as the data distribution is often non-i.i.d.\@ across the vehicles. This paper presents a new confidence-based federated distillation method to improve the performance of federated learning for steering angle prediction. 
Specifically, it proposes the novel use of entropy to determine the \yt{predictive confidence} of each local model, and then selects the most confident local model as the teacher to guide the learning of the global model. 
A comprehensive evaluation of vision-based lane centering shows that the proposed approach can outperform FedAvg and FedDF by 11.3\% and 9\%, respectively. 
\end{abstract}
%
%
\section{Introduction}
\label{sec:introduction}

Autonomous driving is an emerging field that has the potential to evolve the way humans transport. 
A fundamental challenge of autonomous driving is maintaining the vehicle in the center of the lane, e.g., lane centering, by adjusting the steering angle at different driving conditions. 
Recent advancements leverages deep neural networks to predict steering decisions directly from images captured by the car cameras. 
%
Machine learning-based steering angle prediction needs to consider the vehicle's limitation in uploading large amounts of potentially private data for model training. 
Federated learning has the potential to address these limitations by enabling a large number of vehicles to collaboratively train a global model without sharing their private data. Each vehicle shares only its locally trained model with the server, and the server aggregates all the local models to derive the global model.  

\begin{sloppypar}
Conventional federated learning aggregates the local models by performing parameter-level averaging (FedAvg~\cite{mcmahan2016communication}). However, its performance degrades when data is non-independent identically distributed (non-i.i.d.\@) across the clients~\cite{li2020federated, karimireddy2020scaffold, zhao2018federated}. 
Related works have studied regularization, distillation-based methods to mitigate this degradation. 
FedProx~\cite{li2020federated} regularizes the local training with $L_2$ distance between the local and global models, by introducing an additional regularization term in the local training objective function.
FedDF~\cite{lin2020ensemble} aggregates the local models through knowledge distillation (KD), instead of averaging the models, considering each local model as a teacher model and the global model as the student model. 
Conventional KD transfers knowledge from a teacher to a student by minimizing the Kullback–Leibler divergence between the teacher's and student's soft logits (i.e., the softmax output) and the cross entropy loss of the data labels~\cite{hinton2015distilling}.
FedDF~\cite{lin2020ensemble} treats each local model as a teacher. It first calculates the soft logits of each teacher on unlabeled public data, and then uses the averaged soft logits from all the teachers to transfer their knowledge to the student, the global model. 
\yt{Moreover, researchers improved federated distillation methods considering predictive confidence of all the teachers, i.e., confidence-based federated distillation, for classification problems.} FedET~\cite{cho2022heterogeneous} is a softmax-dependent approach, relying on the softmax output to calculate the confidence of each local model. FedAUX~\cite{sattler2021fedaux} relies on a logistic regression classifier to calculate the confidence score.

\end{sloppypar}
These related federated learning methods have their limitations.
FedProx penalizes local updates when local models diverge significantly from the global model, leading to very small local updates and thus very slow convergence~\cite{li2022federated}. 
FedDF neglects the heterogeneity in data distribution, like the conventional federated learning, and cannot fully address the performance issue in highly non-i.i.d.\@ scenarios. Prior work~\cite{zhang2022fine} confirms that the performance gap between FedDF and FedAvg reduces significantly with increasing data heterogeneity. Specifically, the performance gap reduces by 40\% when the data distribution changes from i.i.d.\@ to non-i.i.d.
Moreover, the prediction models for lane centering solve regression problem and do not use softmax; hence the existing softmax-based KD methods do not work for such models. \yitao{Similarly, approaches relying on classification techniques (such as FedAUX) do not work on regression problems either.}

To address these limitations, we propose a new confidence-based federated distillation method to improve the performance of federated learning for steering angle prediction under non-i.i.d.\@ data distribution across the vehicles. We analyze local model divergence and observe that the \textit{entropy of penultimate layer output} shares a similar pattern as the root-mean-square error (RMSE) loss of the model and can serve as a good indicator of the confidence of each local model.
We then propose a novel use of entropy to identify the most confident local model to transfer the knowledge to the global model and improve the performance of federated learning. 
Specifically, in our method, the server performs inference on the received local models using public data. Given a specific sample, it calculates the entropy of each local model's penultimate layer output, and uses the penultimate layer output of the model with the lowest entropy to supervise the training of the global model. In practice, we put back together the penultimate layer output of the best local model for each sample in the batch into a matrix, so we can use it to supervise the training of the global model for the whole batch of samples by minimizing the RMSE between the global and local model output matrices.

We evaluate our proposed approach with comprehensive experiments on a real-world dataset (Udacity~\cite{udacity2018}) using popular deep neural networks (PilotNet~\cite{bojarski2016end} and ResNet-8~\cite{he2016deep}). The experimental results show that our proposed approach outperforms state-of-the-art methods. Specifically, our approach outperforms FedAvg and FedDF in root-mean-square-error (RMSE) by 11.3\% and 9\%, respectively. 

Our main contributions are summarized as follows: 1) \yitao{To the best of our knowledge, we are the first to use confidence-based federated distillation for regression problems such as vision-based lane centering;} 
2) We propose confidence-based federated distillation to address non-i.i.d.\@ data distribution. It leverages the novel use of entropy to determine the model confidence of each local model and select the most confident model to guide the learning of the global model;
3) We provide an extensive comparative evaluation under both i.i.d.\@ and non-i.i.d.\@ settings using representative models and real-world dataset.

\section{Methodology}
\label{sec:methodology}

\begin{figure}[t]
	\centering
	\includegraphics[width=8cm]{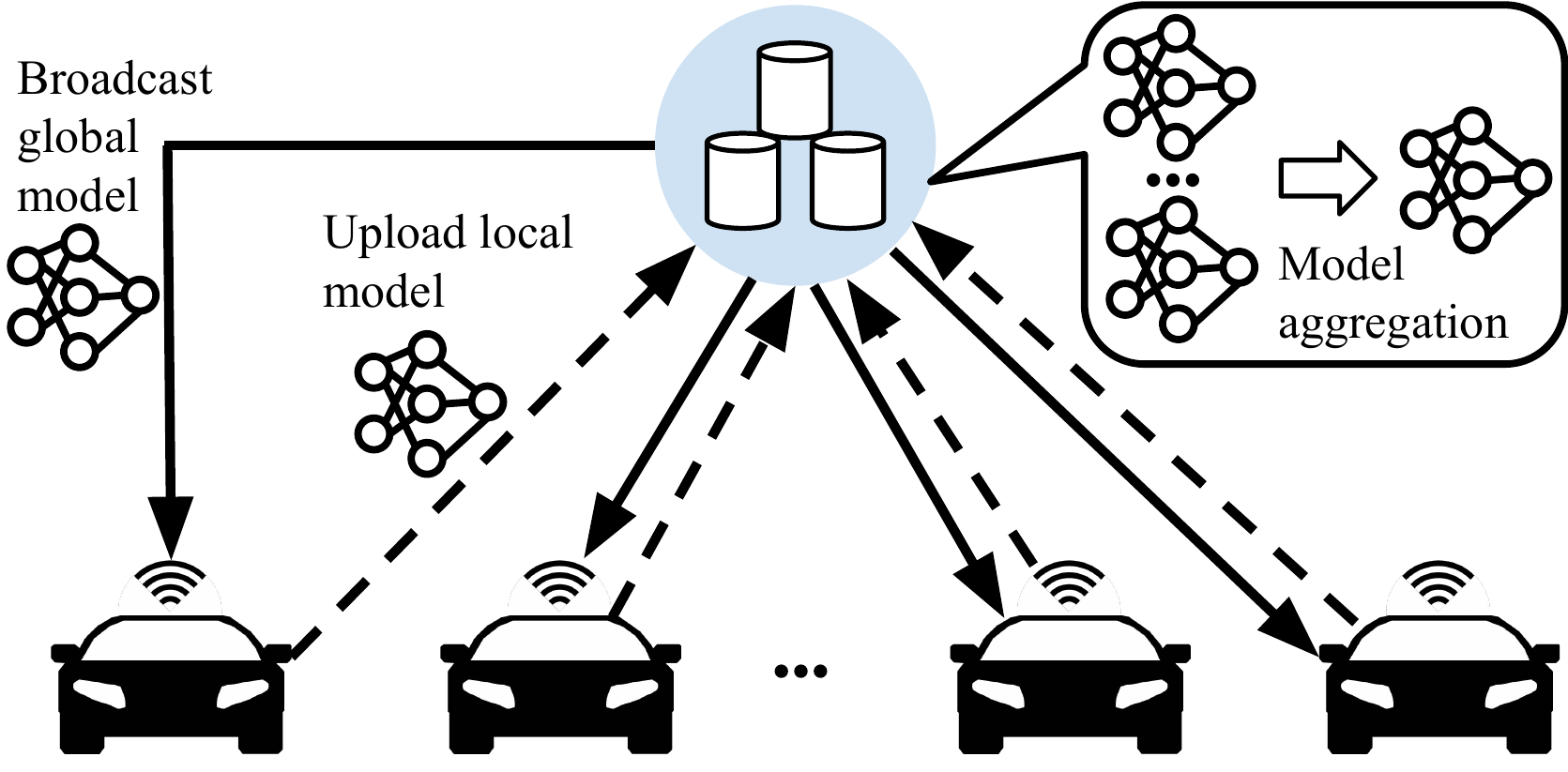}
	\caption{Architecture of federated learning-based vehicle steering angle prediction.}
    \vspace{-10pt}
	\label{fig:fl_cars}
\end{figure}

\subsection{System Architecture}

Fig.~\ref{fig:fl_cars} illustrates our federated learning setup for vision-based lane centering, where multiple vehicles interact with a central server, and each vehicle has access to its local training samples but the central server does not. 
The global model resides in the central server, whereas each local model resides on each vehicle. The federated learning process is iterative. 
Each iteration in this process is a communication round. At the beginning of each communication round, the server broadcasts an initial global model to all the vehicles. The vehicles then copy the received global model as their local models and perform a few local training steps to update their local models. When the local training steps are completed, all vehicles return the trained local models to the server for aggregation.  

\subsection{Confidence-based Federated Distillation}

Previous studies~\cite{karimireddy2020scaffold, zhao2018federated} have shown that non-i.i.d.\@ local data leads to divergence in local models. Such divergence degrades the performance of the global model if it aggregates all the local models indiscriminately. Since the divergence is rooted in the local models' different feature representation abilities, instead of trying to limit the divergence as in the related solutions~\cite{li2020federated, karimireddy2020scaffold}, we ask a different question: can we leverage the divergence to solve the non-i.i.d.\@ local data problem? 

We propose to understand the confidence of each local model in performing the prediction task for a given unlabeled input, and selects the most confident local model as the teacher to guide the training of the global model. 
But this task is non-trivial as techniques to obtain predictive confidence, though rich, cannot directly apply to deep neural networks~\cite{ren2021survey}. For a classification task, one may obtain predictive confidence from softmax output, but prior study has shown that it is unreliable~\cite{wang2016cost}. 
Fortunately, prior arts from active learning provide us the inspirations for solving this problem. 
Active learning attempts to maximize the learning performance with the fewest training samples possible~\cite{ren2021survey}. Learning from only the most uncertain training samples is one of the most popular strategies~\cite {ash2020deep, wang2022boosting}. The key of this strategy is capturing model uncertainty, i.e., how uncertain the model is in its prediction given an unlabeled input~\cite{lewis1994heterogeneous}.

Our goal is then to identify the local model with the least uncertainty, or in another word with the most confidence, given an unlabeled sample. 
RMSE loss of the local model can be a solution to this task, as for a well-trained model, lower the loss, lower the uncertainty, and higher the confidence. But directly computing model loss is infeasible since public data lack ground-truth labels. To address this challenge, we adopt entropy to approximate this loss, inspired by existing effort in active learning~\cite{wang2022boosting}. Without ground-truth information, it is reasonable to use entropy to approximate RMSE loss due to the following two reasons: 1) entropy and RMSE share similarity mathematically~\cite{bush2008computing}; 2) entropy can be used as an effective regularizer in training involving unlabeled data~\cite{berthelot2019mixmatch}. We empirically examine this approximation in Section~\ref{sec: val_entroy}, and the result confirms that entropy shares a similar pattern as the RMSE loss. Therefore, this approximation is reasonable. 
\yt{Considering the constrained computing resource equipped on a vehicle, our approach performs the knowledge distillation on the server side, leaving the local training on the vehicle unaffected.}

Our proposed confidence-based federated distillation approach works as follows. \yt{During model aggregation, given a batch of unlabeled public data, the server performs inference on this batch of data using all the local models.} The server then calculates the entropy of each local model's penultimate layer output, which captures the uncertainty of the prediction. 
We use $H(\cdot)$ to denote the entropy function. The entropy of the penultimate layer output X on sample $j$, $H(X_j)$, can be calculated by
\begin{equation}
\footnotesize
\label{eq: entropy}
\yt{H(X_j) := -\sum_i X_{j,i}log{X_{j,i}}},
\end{equation}
\yt{where $X_{j,i}$ denotes the $i^{th}$ element in the penultimate layer output.} The higher the entropy, the higher the uncertainty, and the lower the confidence. 

After retrieving the entropy of all the local models on this sample, we sort all the entropy, identify the most confident local model as the teacher, and use its penultimate layer output to supervise the training of the global model on this sample.
Our approach uses RMSE as the loss function (denoted by $\ell$), 
\begin{equation}
\footnotesize
\label{eq: loss_func}
    \yt{\ell = \sqrt\frac{\sum_{i=1}^N (y_i^t - y_i^s)^2}{N}},
\end{equation}
\yt{where $y_i^t$ denotes the $i^{th}$ element in the penultimate layer output from the teacher, $y_i^s$ denotes the $i^{th}$ element in the output from the student, and $N$ denotes the number of elements in that output.}
We aim to minimize the RMSE between the penultimate layer output of the teacher and the student by updating the student's weights in the backward pass, using
\begin{equation}
\footnotesize
\label{eq: distill}
w_{r,j}^{s} := \yt{w_{r, j-1}^{s}} - \eta\frac{\partial\ell(f(w_{r, j-1}^{s}, d), f(w_{r, j-1}^{t}, d))}{\partial w_{r,j-1}^{s}}, 
\end{equation}
where 
$w^s_{r,j}$ denotes the student model weights at round $r$, step $j$, $w_{r, j-1}^{t}$ denotes the teacher model weights at round $r$, step $j-1$, $f(\cdot)$ denotes the forward pass of a neural network, $\eta$ represents the learning rate, $\ell$ denotes the RMSE loss, and $d$ denotes a batch of unlabeled data. 

In practice, we put back together the penultimate layer output of the best teacher for each sample in the batch into a matrix, so we can use it to supervise the training of the global model for the whole batch of samples by minimizing the RMSE between the teacher and student output matrices.
Algorithm~\ref{algo: fd} summarizes the confidence-based federated distillation algorithm. 
\begin{algorithm}
\small
\caption{\yt{Confidence-based Federated Distillation. $R$ is the number of training rounds; $|n|$ is the number of vehicular-clients; $T$ is the number of steps in each round.}}
\label{algo: fd}
\begin{algorithmic}[1]
\Procedure{Server}{}
    \For {each communication round $r = 1,..., R$}
        \State $S_t\gets${random subset of} of K vehicular-clients
        \For {each vehicular-client k $\in S_t$ \textbf{in parallel}}
            \State $w_{r}^k\gets$ ClientUpdate$(k, w_{r-1})$ \Comment{Algorithm 2} 
        \EndFor
        \State Initialization for distillation $w_{r}\gets\sum_{k=1}^K \frac{1}{|n|}w_{r-1}^k $
        \For {\yt{$j$ in ${1,...,T}$}} 
            \State mini-batch of unlabeled samples d
            \For {each vehicular-client k $\in S_t$ }
                \State calculate the confidence with~(\ref{eq: entropy})
            \EndFor
            \State Sort teacher models by confidence
            \State Select the most confident teacher
            \State update the server model with~(\ref{eq: distill})  
        \EndFor
        \State \yt{$w_r\gets\ w_{r,T}$}
    \EndFor
    \State \textbf{return} $w_R$
\EndProcedure
\end{algorithmic}
\end{algorithm}
\vspace{-20pt}

\begin{algorithm}
\small
\caption{Client Update in FedAvg~\cite{mcmahan2016communication}. The $K$ vehicular-clients are indexed by $k$; B is the local mini-batch size, E is the number of local epochs, and $\eta$ is the learning rate. } 
\begin{algorithmic}[1]
\Procedure{ClientUpdate}{$k, w_{r-1}^k$}
    \State Client receives $w_{r-1}^k$ from server and copies it as $w_r^k$
    \For {each local epoch $i$ from 1 to $E$}
    \State $\mathfrak{B}$ $\gets$ (split $P_k$ into batches of size $B$) 
        \For {batch $b$ $\in$ $\mathfrak{B}$} 
        \State $w_r^k\gets w_r^k - \eta\triangledown\ell(w_{r-1}^k;b)$
        \EndFor
    \EndFor
    \State \textbf{return} $w_r^k$ to server
\EndProcedure
\end{algorithmic}
\end{algorithm}

\section{Evaluation}
\label{sec:evaluation}

\subsection{Setup}

\textbf{Dataset and Model.}
We perform evaluations on a widely-used dataset, Udacity~\cite{udacity2018}. The dataset consists of 100K images from various weather and driving conditions, recorded from car cameras, from five distinct trips. 
To implement our algorithm, we employ a federated network of five vehicular clients, limited by the dataset. We evaluate our algorithm in both i.i.d.\@ and non-i.i.d.\@ settings. For the i.i.d.\@ setting, we pool all the data together and then randomly and uniformly distribute them among all the vehicular-clients. For the non-i.i.d.\@ setting, we consider the training data from each trip as the local training data to a vehicular-client. 

We evaluate two popular neural networks, PilotNet~\cite{bojarski2016end} and ResNet-8~\cite{he2016deep}. PilotNet consists of five convolutional layers and four fully-connected layers with 559K parameters. ResNet-8 has 74K parameters, consisting of six convolutional layers and two fully-connected layers. 
We consider ResNet-8, motivated by previous success~\cite{valiente2019controlling, du2019self} of using a ResNet-50 with transfer learning techniques. 
But ResNet-50 is too large for resource-constrained vehicles, so we resort to ResNet-8, a reduced version of ResNet-50.

\smallskip
\noindent\textbf{Implementation details.} We train our local models using Adam optimizer with 1e-5 weight decay. The learning rate is 1e-4, and we use a constant learning rate~\cite{mcmahan2016communication}. The batch size is 64 in PilotNets experiments but reduced to 32 in ResNet-8 experiments due to GPU memory constraints. The dataset is partitioned into training/test/public with 7:2:1 as in prior literature~\cite{cho2022heterogeneous}. We train all the models for 100 rounds~\cite{lin2020ensemble}. The local epoch is 5. We consider full-client participation. 

\subsection{Entropy Effectiveness in Capturing Model Divergence} \label{sec: val_entroy}
In this experiment, we evaluate the effectiveness of using entropy to capture the model divergence. As the Udacity dataset consists of data collected from five distinct trips, we partition the data collected in each trip into public and private partitions---10\% is public and 90\% is private. Then the private data of each trip is used to train a model. 
To better explain this experiment's setting and result, we introduce an index $i = [5]\in\{1,..,5\}$ to identify each trip. 
For trip $i$, we denote the private data of this trip as $private_i$, the public data as $public_i$, and the model trained on $private_i$ as $model_i$. 
Next, we perform inference using the models trained on the public data from different trips and calculate the model loss, assuming each model has access to the ground-truth labels. 
We propose two hypotheses: 1) when performing inference on public data, a model trained on the private data from the same trip should have a lower loss value than the models trained on private data from different trips; 2) without access to the ground-truth labels, entropy behaves similarly as the loss. 

\begin{table}[t]
\centering
\footnotesize
\caption{The divergence on loss. Each element represents the average loss value of a model performing inference on public data belonging to a certain trip.}
\vspace{-10pt}
\label{tab:divergence_loss}
\begin{adjustbox}{width=0.47\textwidth}
\begin{tabular}{@{}cccccc@{}}
\toprule
\multicolumn{1}{l}{} & $Public_1$         & $Public_2$         & $Public_3$         & $Public_4$         & $Public_5$         \\ \midrule
$Model_1$            & \textbf{0.005} & 0.252          & 0.194          & 0.432          & 0.060          \\
$Model_2$            & 0.149          & \textbf{0.016} & 0.221          & 0.378          & 0.128          \\
$Model_3$            & 0.114          & 0.172          & \textbf{0.004} & 0.383          & 0.121          \\
$Model_4$            & 0.169          & 0.304          & 0.121          & \textbf{0.011} & 0.269          \\
$Model_5$            & 0.071          & 0.242          & 0.214          & 0.403          & \textbf{0.002} \\ 
\bottomrule
\vspace{-15pt}
\end{tabular}
\end{adjustbox}
\end{table}

\begin{table}[t]
\footnotesize
\centering
\caption{The divergence on entropy. Each element represents the average entropy value of a model performing inference on public data belonging to a certain trip.}
\vspace{-10pt}
\label{tab:divergence_entropy}
\begin{adjustbox}{width=0.47\textwidth}
\begin{tabular}{@{}cccccc@{}}
\toprule
\multicolumn{1}{l}{} & $Public_1$         & $Public_2$         & $Public_3$         & $Public_4$         & $Public_5$         \\ \midrule
$Model_1$            & \textbf{2.573} & 2.532          & 2.625          & 2.544          & 2.643          \\
$Model_2$            & 2.857          & \textbf{1.672} & 2.614          & 2.413          & 2.607          \\
$Model_3$            & 2.763          & 2.703          & \textbf{1.933} & 1.754          & 2.766          \\
$Model_4$            & 2.811          & 2.246          & 2.437          & \textbf{0.113} & 2.478          \\
$Model_5$            & 2.787          & 2.276          & 2.535          & 2.623          & \textbf{2.286} \\ \bottomrule
\vspace{-20pt}
\end{tabular}
\end{adjustbox}
\end{table}

Table~\ref{tab:divergence_loss} shows the RMSE loss results and confirms our first hypothesis. 
Each element in the table represents the averaged loss of each model performing inference on public data belonging to a specific trip. 
The boldfaced elements represent the lowest loss in a column.
The elements along the diagonal are the loss where the public data and private data are from the same trip. 
All the elements along the diagonal are boldfaced, which confirms our hypothesis. 

Table~\ref{tab:divergence_entropy} shows the entropy values in the same setting and confirms our second hypothesis. The entropy values along the diagonal are the lowest among other values in the same column. This observation confirms that entropy can capture the model divergence in the loss, as the entropy results align with the loss results perfectly. Hence it is an effective way to identify the model with the best knowledge.

\subsection{Confidence-based Federated Distillation Effectiveness}

\begin{figure}[t]
	\begin{subfigure}{0.45\columnwidth}
		\centering
		\includegraphics[width=4cm]{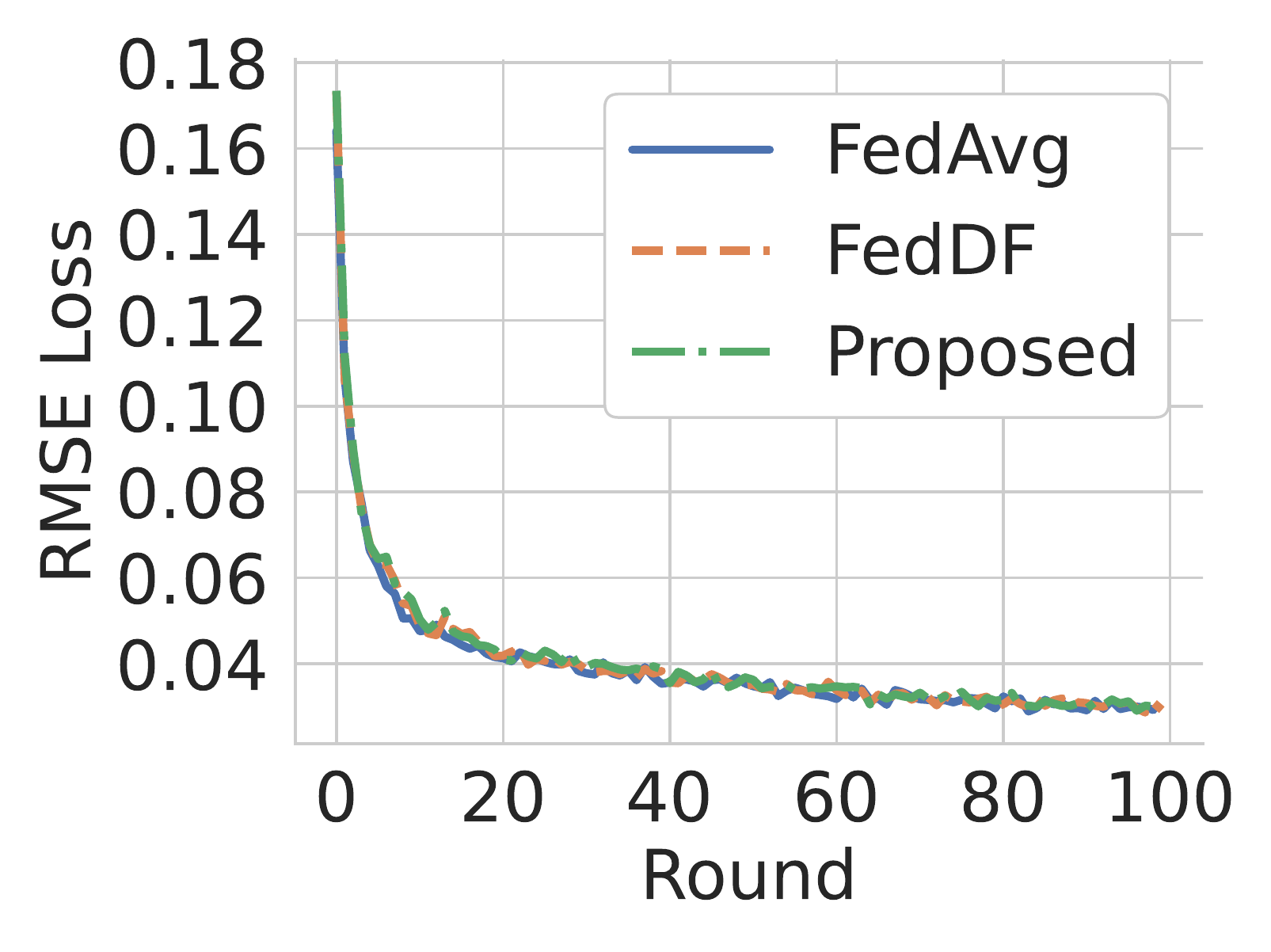}
		\vspace{-15pt}
		\caption{PilotNet i.i.d.}
		\label{pilotnet_iid}
	\end{subfigure}
	\begin{subfigure}{0.45\columnwidth}
		\centering
        \includegraphics[width=4cm]{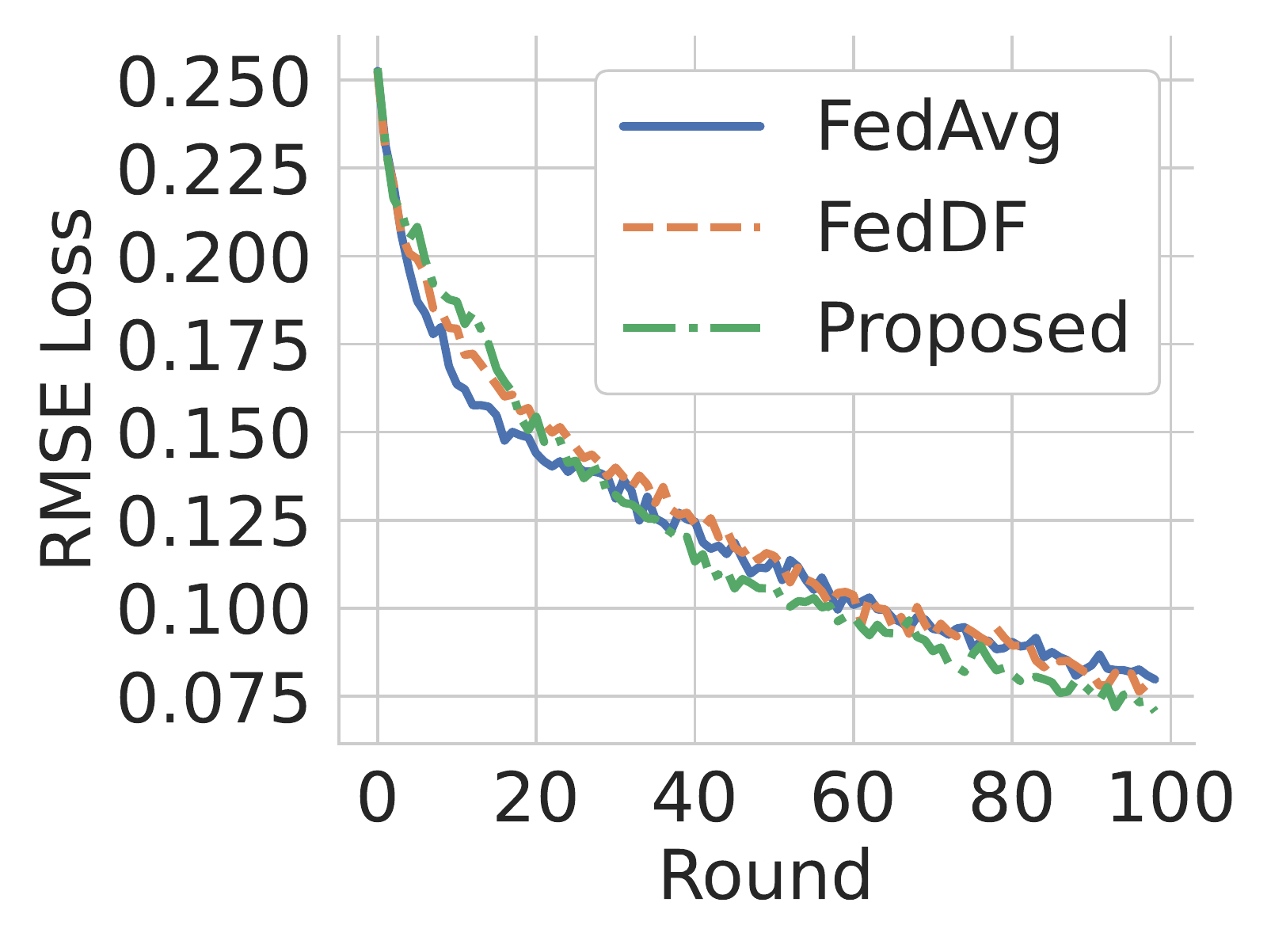}
		\vspace{-15pt}
		\caption{PilotNet non-i.i.d.}
		\label{pilotnet_noniid}
	\end{subfigure}
	\vspace{-5pt}
	\caption{PilotNet results under i.i.d.\@ and non-i.i.d.\@ settings.}
 	\vspace{-10pt}
\end{figure}

Fig.~\ref{pilotnet_iid} shows the RMSE loss of PilotNet under the i.i.d.\@ setting across all communication rounds. 
Our method achieves comparable RMSE compared to the FedAvg and FedDF. Specifically, the RMSE values of FedAvg, FedDF, and our approach are 0.02925, 0.02924, and 0.02967, respectively.
Since the data distribution is i.i.d.\@, we can expect FedAvg to perform well. Our results show that the distillation-based federated learning methods, FedDF and our method deliver the same level of performance. 
Fig.~\ref{pilotnet_noniid} illustrate the RMSE results of PilotNet on handling the more challenging, non-i.i.d.\@ data. Our method outperforms the baselines significantly. Specifically, it achieves a lower RMSE than FedAvg and FedDF, by 11.3\% and 9\%, respectively. Our method leverages the model divergence between all the teacher models and uses the knowledge from the best teacher model to achieve the best learning outcome of the student model. Compared to our method, FedDF treats all the teacher models equally regardless of their data distribution, leading to a worse performance. \yt{Note that our method requires well-trained teacher models to provide accurate confidence values to achieve good results. Hence, during the early training rounds (in the first 30 rounds), our method performs worse than the other two baselines.}

\begin{figure}[t]
	\begin{subfigure}{0.45\columnwidth}
		\centering
		\includegraphics[width=4cm]{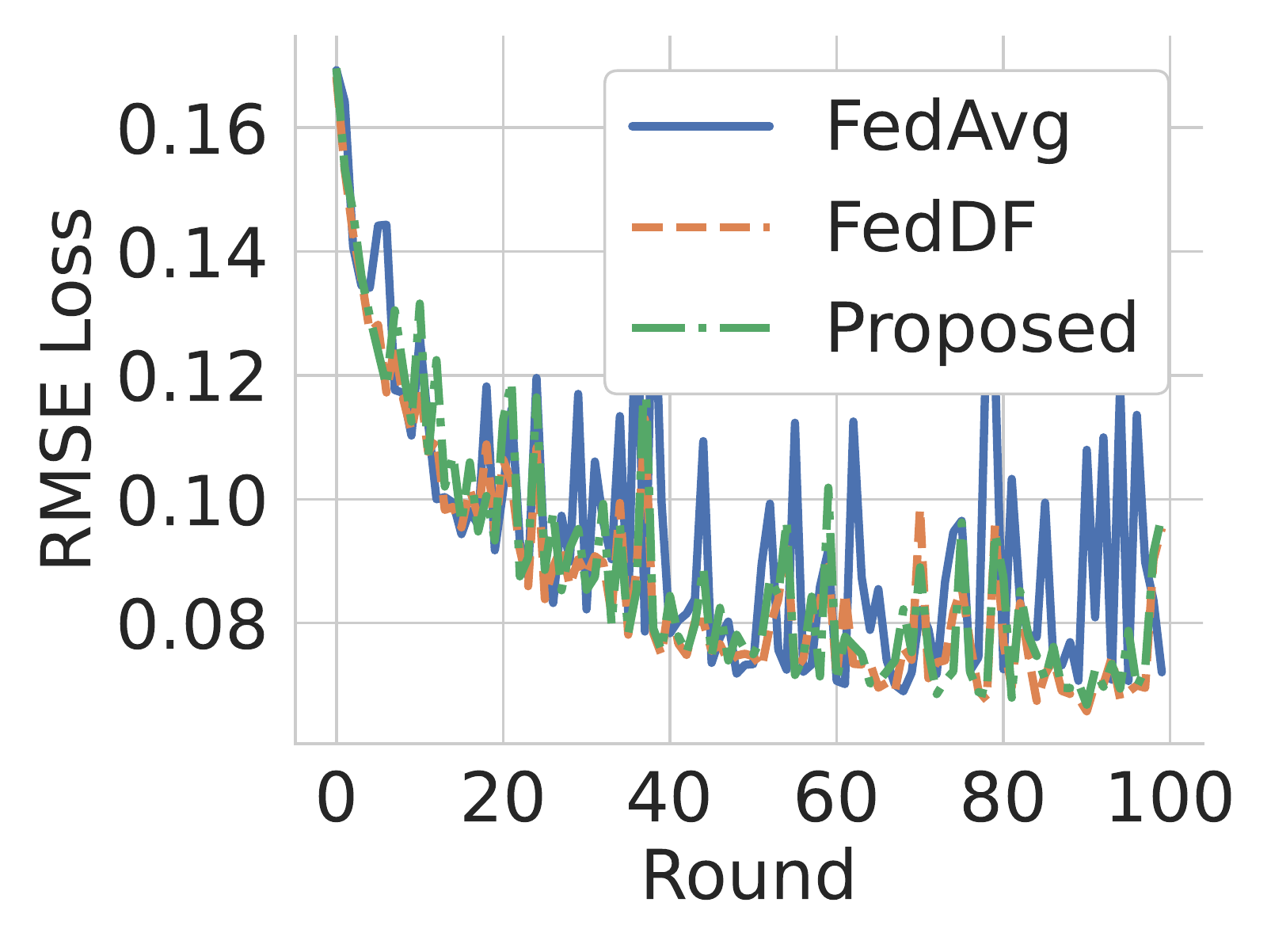}
		\vspace{-15pt}
		\caption{ResNet-8 i.i.d.}
		\label{res8_iid}
	\end{subfigure}
	\begin{subfigure}{0.45\columnwidth}
		\centering
		\includegraphics[width=4cm]{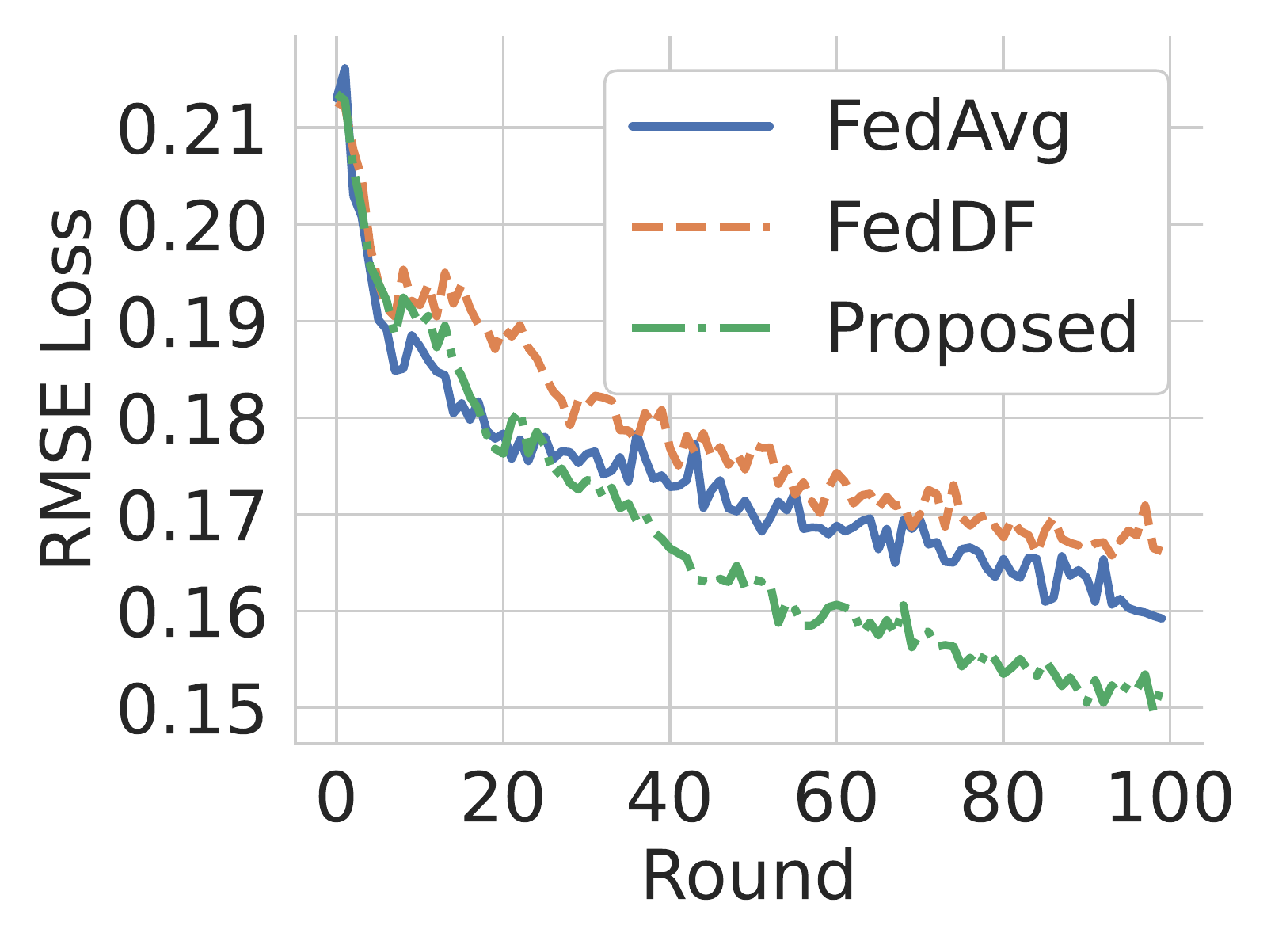}
		\vspace{-15pt}
		\caption{ResNet-8 non-i.i.d.}
		\label{res8_noniid}
	\end{subfigure}
	\vspace{-5pt}
	\caption{ResNet-8 results under i.i.d.\@ and non-i.i.d.\@ settings.}
        \vspace{-10pt}
\end{figure}


Fig.~\ref{res8_iid} illustrates the i.i.d.\@ results of ResNet-8. Similar to the observation from the PilotNet, our method achieves comparable results to the baselines. 
But we observe that the learning curves of ResNet-8 are noisier than PilotNet. More hyperparameter tuning may be required due to the model architecture difference between ResNet-8 and PilotNet.
Fig.~\ref{res8_noniid} illustrates the non-i.i.d.\@ results of ResNet-8. Our method outperforms FedAvg and FedDF baselines by 4.2\% and 8.3\%, respectively, confirming that our method also benefits the ResNet model.

\section{Conclusion and Future Work}
\label{sec:conclusions}

This paper presents confidence-based federated distillation for vision-based lane-centering in autonomous driving to address the common non-i.i.d.\@ data distribution across vehicles and enable effective federated learning for high accuracy, low overhead, and strong privacy. It proposes the novel use of entropy, calculated from the penultimate layer's output, as the indicator of confidence and selects the most confident local models as a teacher to guide the training of the global model. Our evaluation confirms that entropy is indeed a good indicator of model uncertainty. It also shows that our proposed method improves the performance of the global model under non-i.i.d.\@ data distribution compared to state-of-the-art federated learning methods. 
\yt{In our future work, we will conduct more experiments using larger datasets, such as Comma-ai~\cite{santana2016learning} and BDD~\cite{yu2020bdd100k}, and explore the feasibility of applying our proposed confidence-based federated distillation methods to other domains.  }

\bibliographystyle{IEEEbib}
\bibliography{yitao}

\end{document}